\documentclass[]{bytedance}
\usepackage[toc,page,header]{appendix}

\usepackage{minitoc}
\usepackage{amsfonts}
\usepackage{amssymb}
\usepackage{tabularx}
\usepackage{listings}
\usepackage{xcolor}
\usepackage{cancel}

\usepackage{tabulary,multirow,xspace}
\usepackage{fixmath,mathtools,nicefrac,mmstyle}
\usepackage{subcaption}
\usepackage{caption}
\usepackage{wrapfig} 
\usepackage[misc]{ifsym} 
\usepackage{colortbl}

\usepackage{wrapfig}
\usepackage{multicol}
\usepackage[most]{tcolorbox}
\usepackage{pifont}

\definecolor{codegreen}{rgb}{0,0.6,0}
\definecolor{codegray}{rgb}{0.5,0.5,0.5}
\definecolor{codepurple}{rgb}{0.58,0,0.82}
\definecolor{backcolour}{rgb}{0.95,0.95,0.92}
\definecolor{boxblue}{RGB}{57,89,163}
\definecolor{boxbluebg}{RGB}{230,237,250} 

\lstdefinestyle{mystyle}{
    backgroundcolor=\color{backcolour},   
    commentstyle=\color{codegreen},
    keywordstyle=\color{magenta},
    numberstyle=\tiny\color{codegray},
    stringstyle=\color{codepurple},
    basicstyle=\ttfamily\footnotesize,
    breakatwhitespace=false,         
    breaklines=true,                 
    captionpos=b,                    
    keepspaces=true,                 
    numbers=none,                    
    numbersep=5pt,                  
    showspaces=false,                
    showstringspaces=false,
    showtabs=false,                  
    tabsize=2
}
\definecolor{mygray1}{gray}{.95}
\definecolor{mygray2}{gray}{.9}
\definecolor{mygray3}{gray}{.95}
\usepackage{pifont}

\newlength\savewidth
\newcolumntype{x}[1]{>{\centering\arraybackslash}p{#1pt}}

\newcommand{\app}{\raise.17ex\hbox{$\scriptstyle\sim$}}

\usepackage{xcolor}
\usepackage{graphicx}
\usepackage{amssymb}
\usepackage{pifont}
\usepackage{floatrow}
\usepackage{amsmath} 
\usepackage{float}
\usepackage{wrapfig}
\usepackage{multirow}
\usepackage{tcolorbox}
\tcbuselibrary{breakable, skins, raster}
\usepackage{listings}
\usepackage{listings}

\usepackage{algorithm}
\usepackage{algorithmic}
\definecolor{myblue}{RGB}{210, 225, 255}
\definecolor{mytextblue}{RGB}{51, 161, 201}
\definecolor{mypurple}{RGB}{218, 112, 214}

\definecolor{commentgreen}{rgb}{0.1, 0.4, 0.1}
\definecolor{keywordblue}{rgb}{0.1, 0.1, 0.7}
\definecolor{stringred}{rgb}{0.7, 0.1, 0.1}

\lstdefinestyle{mystyle}{
    commentstyle=\color{commentgreen},
    keywordstyle=\color{keywordblue},   
    stringstyle=\color{stringred},
    basicstyle=\ttfamily\scriptsize, 
    breaklines=true,
    keepspaces=true,
    showstringspaces=false,
    frame=none,                     
    language=Python, 
}

\newcommand{\name}{NativeTok}
\title{\name{}:  Native Visual Tokenization for Improved Image Generation}

\author{
\centerline{
Bin Wu \quad 
Mengqi Huang$^*$  \quad  
Weinan Jia \quad 
Zhendong Mao \quad
} 
\texttt{\small \{lilimomotion,jiawn\}@mail.ustc.edu.cn,}
\texttt{\small \{zdmao,huangmq\}@ustc.edu.cn}
}

\affiliation[]{University of Science and Technology of China}

\footnotetext{* Corresponding author.}

\abstract{
VQ-based image generation typically follows a two-stage pipeline: a tokenizer encodes images into discrete tokens, and a generative model learns their dependencies for reconstruction. However, improved tokenization in the first stage does not necessarily enhance the second-stage generation, as existing methods fail to constrain token dependencies. This mismatch forces the generative model to learn from unordered distributions, leading to bias and weak coherence. To address this, we propose \textbf{\textit{native visual tokenization}}, which enforces causal dependencies during tokenization. Building on this idea, we introduce \textbf{NativeTok}, a framework that achieves efficient reconstruction while embedding relational constraints within token sequences. NativeTok consists of: (1) a \textbf{Meta Image Transformer (MIT)} for latent image modeling, and (2) a \textbf{Mixture of Causal Expert Transformer (MoCET)}, where each lightweight expert block generates a single token conditioned on prior tokens and latent features. We further design a Hierarchical Native Training strategy that updates only new expert blocks, ensuring training efficiency. Extensive experiments demonstrate the effectiveness of NativeTok.
}
\checkdata[Project Page]{\url{https://github.com/wangbei1/Nativetok}}

\begin{document}
\maketitle

\section{Introduction}
\label{sec:intro}

Over the past few years, the vision community has witnessed remarkable progress in deep generative models, such as diffusion models\cite{gao2023masked,rombach2022high,peebles2023scalable,podell2023sdxl,ho2020denoising} and autoregressive models\cite{ding2021cogview,esser2021taming,gu2022vector,yu2022scaling,van2017neural,chang2022maskgit,li2023mage,sun2024autoregressive}, elevating image generation quality to unprecedented levels. 
Inspired by the success of large language models (LLMs), large visual models have recently attracted increasing research interest, primarily due to their compatibility with established innovations from LLMs (\textit{e.g.}, scaling laws) and their potential to unify language and vision towards general artificial intelligence (AGI).
Different from the natural tokenization for language such as BPE (Byte Pair Encoding)\cite{shibata1999byte,bostrom2020byte}, the visual tokenization, \textit{i.e.}, converting images into discrete tokens analogous to those used in language models, remains a longstanding challenge for VQ-Based large visual models.

\begin{figure}[ht]
  \centering
  \includegraphics[width=\linewidth]{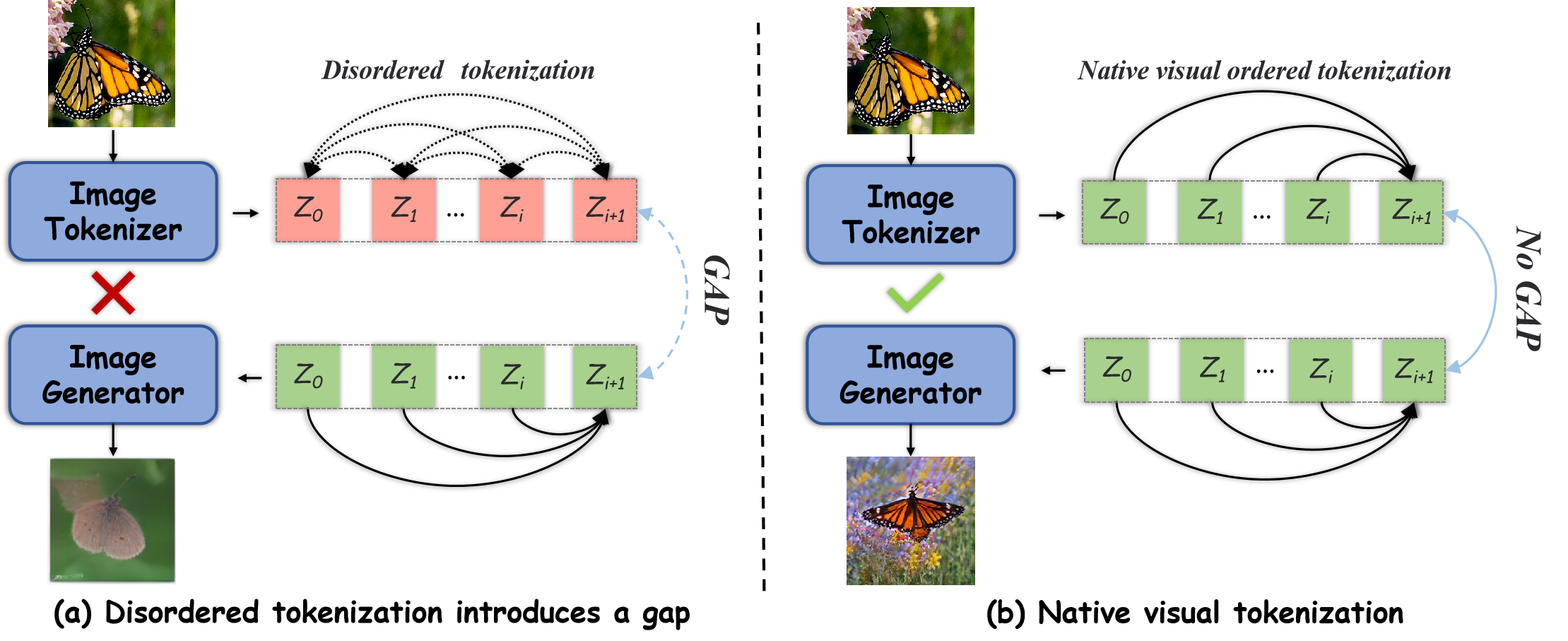}
  \vspace{-4mm}
  \caption{Illustration of our motivation.
  (a) Existing \textit{disordered tokenization} overlooks the essential requirement of relational modeling in the generation stage, as it fails to introduce any token dependency constraints during tokenization, resulting in a gap between the two stages of image generation.
  (b) Our approach proposes \textbf{\textit{native visual tokenization}}, which considers not only reconstruction quality but also imposes relational constraints during tokenization, thereby coupling the two stages of image generation.}
  \label{fig:motivation}
  \vspace{-2mm}
\end{figure}

Most modern large visual models follow a two-stage paradigm, \textit{i.e.}, (1) the first \textit{tokenization stage} learns an image tokenizer by reconstructing images, converting high-dimensional pixels into compressed discrete visual tokens, and (2) the second \textit{generation stage} trains a generative model to model the token distribution via next-token prediction.

Recent works have focused on designing improved image tokenizers with higher compression ratios to accelerate generation process, or higher reconstruction quality to enhance generation quality.
For example, DQVAE \cite{Huang_2023_CVPR} introduces a more compact tokenizer by dynamically allocating varying numbers of tokens to image regions based on their information density.
VAR \cite{tian2025visual} proposes a multi-scale tokenizer combined with a coarse-to-fine "next-scale prediction" generation strategy.
TiTok \cite{yu2025image} significantly accelerates generation by compressing images into shorter one-dimensional serialized representations.
In summary, existing methods primarily focus on first independently designing a better tokenizer, and then training a generative model to capture the distribution of the resulting visual tokens.

However, we argue that existing methods overlook the intrinsic dependency between the tokenization and generation stages, optimizing each stage with separate objectives. This leads to a fundamental misalignment between the disordered tokenization outputs and the structured dependency modeling required during generation.
Specifically, as illustrated in Fig.\ref{fig:motivation}(a), the first stage typically relies on reconstruction loss for supervision during tokenization, without imposing constraints on the intrinsic relationships among tokens, while the second stage essentially models the intrinsic distributional relationships among the tokens learned in the first stage.
This misalignment prompts a critical inquiry: \textbf{\textit{How can a generative model accurately learn visual token distributions if the tokens themselves are inherently disordered in nature?}}

\begin{figure*}[ht]
  \centering
  \includegraphics[width=1\linewidth]{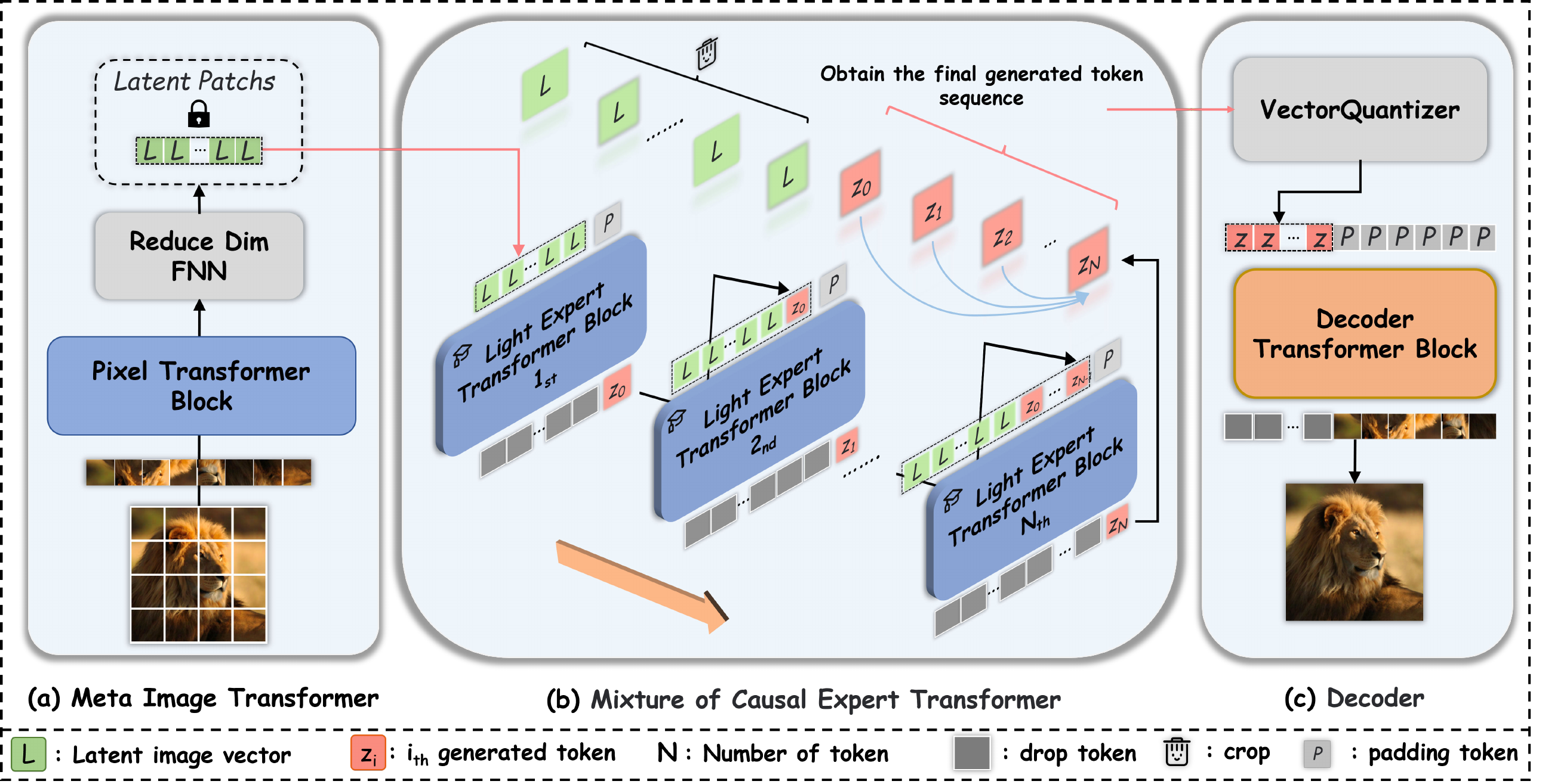}
  \caption{The overview of our NativeTok framework. (a) In Meta Image Transformer (MIT), the image information is initially modeled by a Pixel Transformer, then compresses the image into the latent space. During the subsequent generation process, the latent space information of the image remains locked.  (b) During the token sequence generation process, we define Mixture of Causal Expert Transformer (MoCET). The $i^{th}$ expert transformer block is responsible for generating the $i^{th}$ token. In each generation step, we concatenate the locked latent space image information, all previously generated tokens, and the current mask token, and feed them into the corresponding expert transformer. Once a token is generated, it remains fixed. (c) Once all tokens are generated, they are fed into the decoder to reconstruct the image.}
  \label{fig:framework}
\end{figure*}

As a result, the generation stage can only learn a biased and incomplete distribution from these disordered visual tokens, leading to sub-optimal generative performance.
Moreover, a persistent gap exists in current two-stage generation paradigm, \textit{i.e.}, better tokenization performance in the first stage does not necessarily result in improved generation in the second stage, as it may lead to more complex token interrelationships that further violates the autoregressive principle.

To address the above challenge, we introduce the concept of \textit{native visual tokenization}, which tokenizes images in a native visual order that inherently aligns with the subsequent generation process, leading to more effective and accurate visual token modeling, as illustrated in  Fig.\ref{fig:motivation}(b). 
\textbf{We argue that visual information in an image also follows a global causal order. Analogous to human perception, when observing an image, we tend to first recognize its primary structural components, followed by finer details such as textures. }This observation suggests that the process of visual understanding is inherently autoregressive in nature.
Built upon this concept, we propose \textbf{NativeTok}, a novel visual tokenization framework that disentangles \textit{visual context modeling} from \textit{visual dependency modeling}, thereby enabling the joint optimization of image reconstruction quality and token ordering without conflict. As a result, NativeTok is capable of producing a native visually ordered token sequence directly during the tokenization stage.
Specifically, NativeTok comprises three components: (1) the \textit{Meta Image Transformer (MIT)}, which models image patch features with high-dimension for effective visual context modeling, followed by a dimension switcher that compresses the context output for subsequent efficient visual dependency modeling.
(2) The \textit{Mixture of Causal Expert Transformer (MoCET)} features a novel ensemble of lightweight expert transformer blocks, each tailored to a specific token position. This position-specific specialization allows MoCET to model causal dependencies with greater precision, yielding an ordered token representation.
(3) Finally, the ordered token representation is quantized into discrete tokens and subsequently decoded by a transformer-based decoder.
Moreover, we further propose a novel \textit{Hierarchical Native Training} strategy to efficiently train NativeTok across varying tokenization lengths.
NativeTok not only accomplishes the reconstruction task of the first stage but also explicitly constrains the intrinsic relationships within the token sequence, making it easier for the second-stage generation process to capture and model these dependencies, thereby achieving a close integration between the two stages.
\section{Related Work}
\label{sec:Related Work}

\subsection{Image Tokenization}

Image Tokenization is a fundamental and crucial step in image generation. Image tokenization involves using autoencoders to compress images from high-dimensional, high-resolution space into a low-dimensional, low-resolution latent space, and then using a decoder to reconstruct the image from this latent space. CNN-based autoencoder tokenizers, such as VQ-VAEs \cite{van2017neural,razavi2019generating} ,encode images to a discrete representation, forming a codebook of image information. DQ-VAE \cite{Huang_2023_CVPR} focuses on the information density of different image regions and encodes images using information-density-based variable-length coding. Transformer-based autoencoder tokenizers, such as ViT-VQGAN \cite{yu2021vector} and Efficient-VQGAN \cite{cao2023efficient}, also achieve excellent reconstruction results through the use of transformers. Titok \cite{yu2025image} and MAETok\cite{chen2025masked} achieves excellent reconstruction results by compressing images into one-dimensional sequences with a high compression ratio. Many works further explore improvements in the vector quantization step, such as SoftVQ\cite{chen2024softvq}, MoVQ\cite{zheng2022movq}, and RQ-VAE\cite{lee2022autoregressive}, while  FSQ \cite{mentzer2023finite} propose LFQ methods. In recent work, FlexTok~\cite{bachmann2025flextok} incorporates characteristics of continuous VAEs, utilizing flexible-length 1D encodings to enable a visually ordered reconstruction process, while GigaTok~\cite{xiong2025gigatok} also achieve wonderful generation results. However, these works completely neglects the dependency relationships among tokens that need to be modeled in the second stage, resulting in a fundamental discrepancy between the token dependencies established in the first stage and the image modeling approach in the second stage and do not treat the two stages as a unified process. Therefore, we propose NativeTok, which introduces strong constraints when modeling token dependencies to capture more accurate intrinsic relationships. This enables the generator to more easily learn such associations during the generation process, thereby achieving a true integration of the tokenization and generation stages.

\subsection{Image Generation}

Image Generation typically learns and generates images from the image latent space generated in the image tokenization stage, which enhances generation efficiency. Existing generation methods primarily include three paradigms: GANs\cite{karras2019style,gao2019progan,goodfellow2014generative,karras2020analyzing}, diffusion models\cite{gao2023masked,rombach2022high,peebles2023scalable,podell2023sdxl,ho2020denoising}, and autoregressive models\cite{ding2021cogview,esser2021taming,gu2022vector,yu2022scaling,van2017neural,chang2022maskgit,li2023mage,sun2024autoregressive}. These methods predict the next token or group of tokens, based on conditions and previously generated tokens, until the entire token sequence is generated. However, previous image tokenization approaches fail to consider the inherent requirement of the second-stage generator to model the internal dependencies among tokens. As a result, the generated unordered sequence can only lead to a biased and inherently inaccurate model.
\section{Methodology}
\label{sec:Methodology}
\subsection{NativeTok framework}

To achieve the goal of \textit{native visual tokenization} and ensure tight integration between the tokenization and generation stages, we design the \textbf{NativeTok} framework, as illustrated in Fig.~\ref{fig:framework}. As previously discussed, our aim is to model the ordered relationships among tokens during the tokenization stage. This requires the first-stage tokenization process to ultimately produce an ordered, causally token sequence that aligns with the native visual order:

\begin{equation}
    \boldsymbol z_i = \boldsymbol{\text{Encoder}(X,z_0,z_1, ..., z_{i-1}}).
    \label{eq:reconstruction}
\end{equation}
The modeling of token $z_i$ depends only on the previous tokens $\boldsymbol z_0$ to $\boldsymbol z_{i-1}$ and the image information $\boldsymbol X$, and is independent of any subsequent $\boldsymbol z_{j\space (j>i)}$.

To achieve such a generation process, a naive approach is to apply a causal mask~\cite{vaswani2017attention}. However, we find that this method is ineffective in practice, as the transformer must simultaneously perform image self-modeling and token generation, making it difficult to generate tokens sequentially. Moreover, its performance heavily depends on the capacity of the original model and does not yield substantial improvements, as confirmed by the experimental results presented in the subsequent ablation studies.

Therefore, we propose a novel framework that adopts a divide-and-conquer design. It decouples complex context modeling of image content from the dependency modeling among image tokens. In the image context modeling stage, we employ bidirectional attention to capture global representations of the image. In the token dependency modeling stage, we introduce native visual constraints for token-level dependency modeling, while still allowing each token to adaptively attend to the full image context. 

Specifically, the NativeTok framework consists of the following two components:

\textbf{Meta Image Transformer (MIT):} Instead of directly using the original image embedding for token modeling, we first model the input image $\boldsymbol X$ through the Meta Image Transformer $MIT$, which consists of a series of transformers, followed by a fully connected network (FNN) acting as a dimension switcher to reduce dimensionality, ultimately obtaining $\boldsymbol{X}_{\text{latent}}$, where $\boldsymbol{X}_{\text{latent}}$ represents the rich contextual information of the image in the latent space.
\begin{equation}
    \boldsymbol{X}_{\text{latent}} = \text{FNN}(\text{MIT}(\boldsymbol{X})).
    \label{eq:reconstruction}
\end{equation}

\textbf{Mixture of Causal Expert Transformer (MoCET):} Each token used for representation is modeled by a separate expert transformer. Specifically, our generation process is as follows: (1) Lock $\boldsymbol{X}_{\text{Latent}}$, keeping it unchanged during the generation process to ensure that each subsequent token generation focuses on the same image information. (2) we define an ordered lightweight expert transformer sequence 
       \begin{equation}
        \mathbb T = {\{ \mathbb{T}_0, \mathbb{T}_1, \dots, \mathbb{T}_L \}},
        \label{eq:reconstruction}
    \end{equation} of the same length $L$ as the token sequence. The $i^{th}$ expert transformer block of $\mathbb{T}$ is solely responsible for generating the $i^{th}$ token. (3) In the generation step of each token, we concatenate the locked latent space image information, all previously generated tokens, and the current padding token, and feed them into the corresponding transformer. We retain only a single vector from the original padding token positions as the currently generated token. For example, in the generation process of the $i^{th}$ token:
    \begin{equation}
    \boldsymbol{z}_i = \mathbb{T}_i(\boldsymbol{X}_{\text{latent}},\ z_0, z_1, \dots, z_{i-1},\ z_{\text{padding}}),
    \label{eq:reconstruction}
\end{equation}

\noindent where $\mathbb{T}_i$ denotes the expert transformer block responsible for modeling the $i^{\text{th}}$ token in the Mixture of Causal Experts $\mathbb{T}$. The tokens $\boldsymbol{z}_j\ (j<i)$ represent those that have already been generated, and $\boldsymbol{z}_i$ refers to the current token modeled at the position of $z_{\text{padding}}$. Once $\boldsymbol{z}_i$ is generated, it is fixed and incorporated into the modeling of subsequent tokens, and the process continues until the entire token sequence is produced.

The divide-and-conquer design of NativeTok enables the modeling of visual token dependencies while allowing each image token to adaptively select from the global image context. This strategy aligns well with the previously discussed human perception-inspired modeling paradigm, which progresses from global structures to finer details.

\textbf{Image generation:} To comprehensively evaluate the impact of explicitly introducing constraints during the tokenization stage, we conduct experiments using both autoregressive (AR) and MaskGIT-style generation frameworks. For the AR setting, we adopt LlamaGen~\cite{dosovitskiy2020image} as the generator, while for the MaskGIT setting, we follow recent work~\cite{chang2022maskgit,bao2023all}.

\subsection{Training: Hierarchical Native Training strategy}
\label{sec:sub3.4}
\begin{wrapfigure}{r}{0.5\linewidth}
  \centering
  \includegraphics[width=\linewidth]{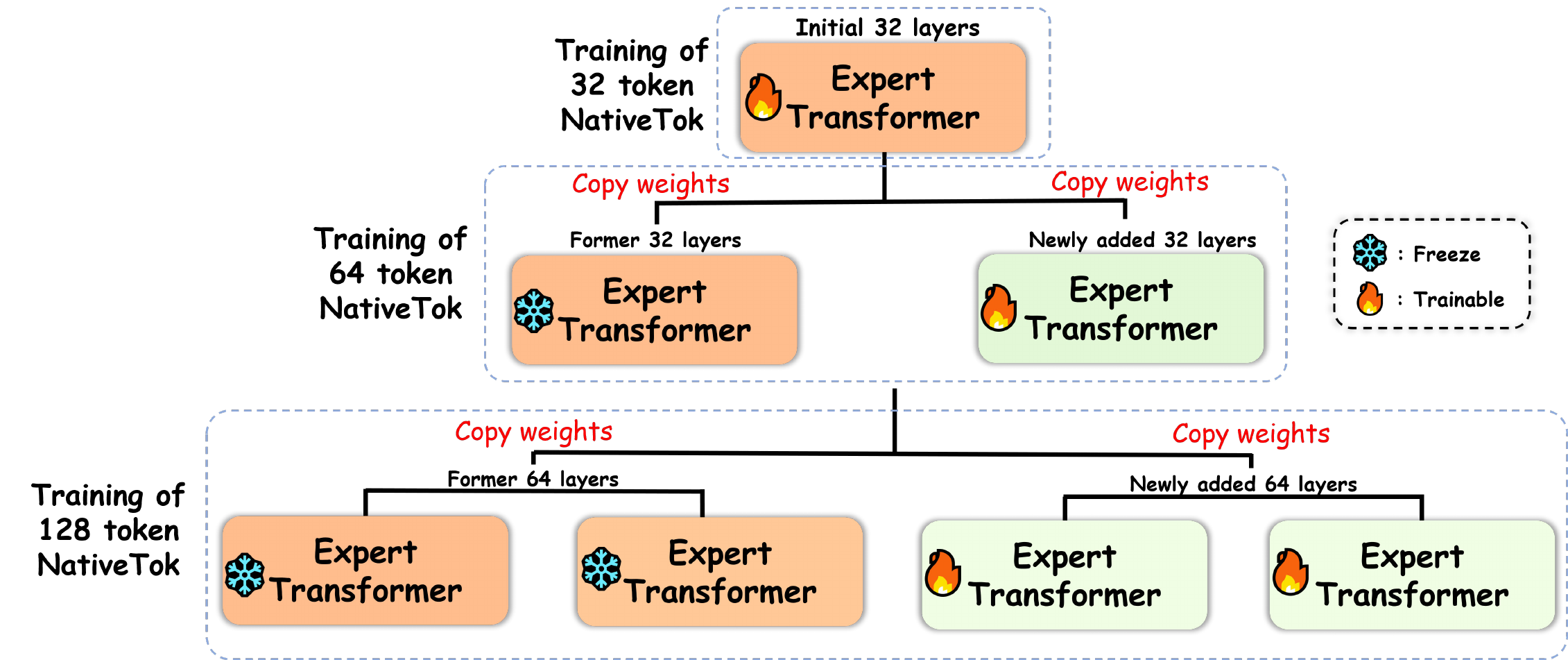}
  \caption{\textbf{Hierarchical Native Training}: We freeze the Meta Image Transformer and existing experts, training only newly added experts initialized with reused weights. This reduces training costs while ensuring the new experts inherit prior modeling capabilities.}
  \label{fig:training}
\end{wrapfigure}

In MoCET, the number of expert layers scales with the token count, as each new token requires an additional expert block. While this design yields strong reconstruction, it also increases training costs. To mitigate this, based on Titok's\cite{yu2025image} two-stage training approach, we propose the \textbf{Hierarchical Native Training} strategy.  

As shown in Fig.~\ref{fig:training}, we first train a 32-token NativeTok with full parameters. For the 64-token version, we reuse the Meta Image Transformer and the first 32 experts, duplicating their weights into the new experts. We then freeze the reused modules and train only the newly added 32 experts and a decoder, reducing trainable parameters to 56\%. The same procedure applies to the 128-token model, where only the additional experts are trained. To further boost reconstruction quality, we freeze parameters for 90\% of the steps and perform full fine-tuning in the last 10\%.

\section{Experiments}
\label{sec:Experiments}
\subsection{Experimental Setups}
\label{sec:sub4.1}

We evaluate our method on class-conditional ImageNet-1K\cite{deng2009imagenet} with $256 \times 256$ image resolution.
The standard Fréchet Inception Distance (FID)\cite{heusel2017gans} is adopted for evaluating the reconstruction and generation quality (denoted as rFID and gFID). rFID is calculated over the entire validation set. gFID follows ADM \cite{dhariwal2021diffusion} by generating 50K samples for FID evaluation in the context of image generation.

\subsection{Main Experiment}

To evaluate the impact of \textit{native visual tokenization} on generation quality, we conduct experiments under both autoregressive (AR) and MaskGIT-style paradigms. For AR generation, we compare NativeTok$_{32}$ against TiTok-L-32 \cite{yu2025image} and VQGAN, using LlamaGen-B as the generator. As shown in Table~\ref{Quantitative_results_1}, our method achieves a 0.23 improvement in gFID over VQGAN. When the final token sequence length is fixed at 32, NativeTok significantly outperforms TiTok-L-32 \cite{yu2025image}, reducing the gFID from 7.45 to 5.23. Notably, despite a slightly worse reconstruction metric (rFID = 2.57), NativeTok achieves better generation performance under the same AR generator.
\begin{figure}[H]
  \centering
  \includegraphics[width=1\linewidth]{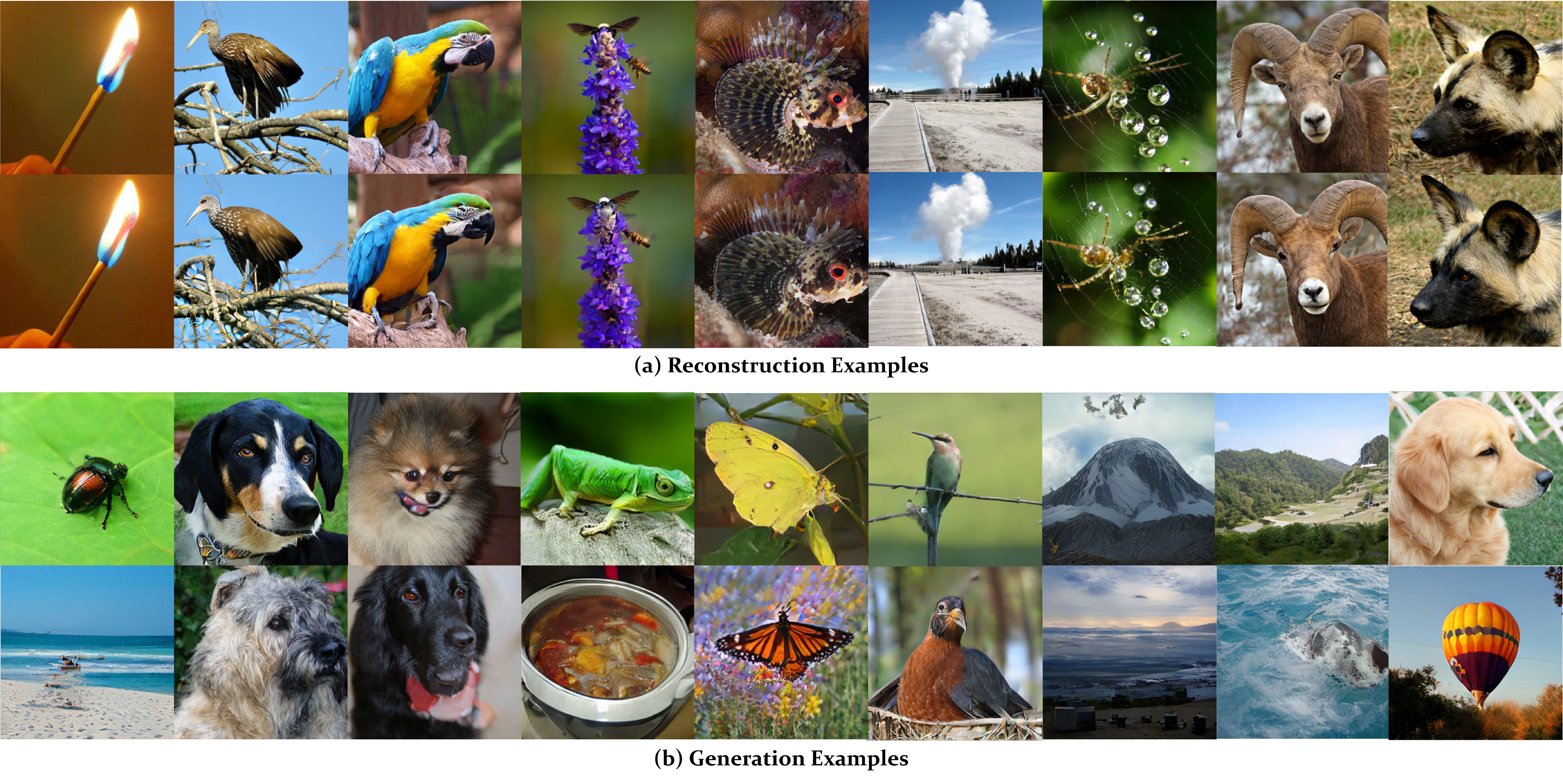}
  \caption{In (a), the top and bottom rows correspond to the original images and their reconstructed results. In (b), we showcase examples of generated images.}
  \label{examples}
\end{figure}
\begin{table}[H]
    \centering
    \resizebox{8cm}{!}{ 
        \begin{tabular}{c c c c c}
            \toprule
            \textbf{Tokenizer} & \textbf{\#Params} & \textbf{\#Tokens} & \textbf{rFID $\downarrow$} & \textbf{(LlamaGen-B) gFID $\downarrow$} \\
            \midrule
            VQGAN & 72M  & 256  & 2.19  & 5.46  \\ 
            TiTok-L-32 & 641M  & 32 & 2.21 & 7.45 \\
            NativeTok$_{32}$ & 616M  & 32 & 2.57  & 5.23 \\ 
            \midrule
        \end{tabular}
    }
    \caption{Main comparison of AR generation performance. \textit{Params} denotes the number of parameters in the tokenizer. \textit{Tokens} denotes the length of the token sequence produced during the tokenization stage.
 All gFID scores are reported using LlamaGen-B as the generator.}
    \label{Quantitative_results_1}
\end{table}
In Table~\ref{main}, we further compare the best-performing version of NativeTok,which is NativeTok$_{128}$ within the MaskGIT framework, using MaskGIT-UVit-L~\cite{chang2022maskgit,bao2023all} as the generator, with several state-of-the-art methods. 
 Under this generation paradigm, NativeTok achieves a gFID of 2.16 with only 287M params, demonstrating strong generation performance.
\begin{table}[h]
\centering
\renewcommand{\arraystretch}{0.70}
\footnotesize
\begin{tabular}{llcc!{\color{gray}\vrule width 0.5pt}clccc}
\toprule
\textbf{Tokenizer} & \#\textbf{Tokens} & \textbf{Codebook Size} & \textbf{rFID$\downarrow$} &
\textbf{Generator} & \textbf{gFID$\downarrow$} & \textbf{P$\downarrow$} & \textbf{Gen-Type} \\
\midrule
\multicolumn{8}{c}{\textit{diffusion-based generative models}} \\

VAE & $1024 \times 4$ & - & 0.62 & UViT-L/2 & 3.40 & 287M & Diffusion \\
& & & & DiT-XL/2 & 2.27 & 675M & Diffusion \\
& & & & SiT-XL/2 & 2.06 & 675M & Diffusion \\
\midrule
\multicolumn{8}{c}{\textit{transformer-based generative models}} \\

RQ-VAE & 256 & 16384 & 3.20 & RQ-Transformer & 4.45 & 1.4B & AR \\
MaskGIT-VQGAN & 256 & 1024 & 2.28 & MaskGIT-ViT & 7.55 & 3.8B & Mask \\
ViT-VQGAN & 1024 & 8192 & 1.28 & VIM-Large & 4.17 & 1.7B & AR \\

VQGAN & 256 & 16384 & 2.19 & LlamaGen-B & 5.46 & 111M & AR \\

& & & & LlamaGen-L & 4.21 & 343M & AR \\
GigaTok-S-S & 256 & 16384 &  1.01 & LlamaGen-B & 4.05 & 111M & AR \\

TiTok-S-128 & 128 & 4096 & 1.71 & MaskGIT-UVit-L & 2.50 & 287M & Mask \\
 &  &  &  & MaskGIT-UVit-L & 1.97 & 287M & Mask \\

VAR & 680 & >4096 & 0.90 & VAR transformer & 1.92 & 2B & AR \\
MAR-KL-16 & 256 & - & 1.22 & MAR-H & 2.35 & 943M & AR \\
MAGVIT-v2 & 256 & 16384 & 1.39 & Open-MAGVIT2-AR-L & 2.51 & 804M & AR \\
FlexTok(d18-d18) & 256 & 64000 & 1.61 & AR Transformers & 2.02 & 1.33B & AR \\
\midrule
\multicolumn{8}{c}{\textit{Ours}} \\
NativeTok$_{128}$ & 128 & 4096 & 1.19 & MaskGIT-UVit-L & 2.16 & 287M & Mask \\
\bottomrule
\end{tabular}
\caption{\textbf{Comparison of Tokenizers and Generators.} In the table, \textbf{P} denotes the number of model parameters, and \textbf{Gen.Type} denotes the generative model type. Specifically, for TiTok-S-128, 2.50 and 1.97 represent the results of 8 and 64 sampling steps, respectively, while the 2.16 result for NativeTok$_{128}$ is obtained with 8 sampling steps.}
\label{main}
\end{table}
This set of experiments confirms that the ordered token sequences generated by NativeTok consistently improve generation quality in both AR and MaskGIT-style settings. \textbf{Since the second-stage generative model essentially learns the distribution established during tokenization, introducing token-level constraints enables more structured and learnable dependencies. }
Examples of reconstruction and generation results are presented in Fig.~\ref{examples}.

\subsection{Ablations}
\begin{table*}[ht]
\centering
\captionsetup{skip=4pt}
\begin{subtable}[t]{0.32\textwidth}
\setlength{\tabcolsep}{1pt}
\renewcommand{\arraystretch}{1}
{\scriptsize
\begin{tabular*}{\linewidth}{@{\extracolsep{\fill}}lcc@{}}
\toprule
\textbf{Model} & \textbf{\#Params} & \textbf{FID$\downarrow$} \\
\midrule
Titok$_{L-32}$ & 641M & 12.99 \\
Titok$_{L(mask)-32}$ & 641M & 12.95 \\
NativeTok$_{32}$ & 616M & 11.19 \\
\bottomrule
\end{tabular*}

}
\caption{Different Attention Mechanisms.}

\label{consolidated_table}
\end{subtable}
\hfill
\begin{subtable}[t]{0.35\textwidth}
\setlength{\tabcolsep}{1pt}
\renewcommand{\arraystretch}{1}
{\scriptsize
\begin{tabular*}{\linewidth}{@{\extracolsep{\fill}}lccc@{}}
\toprule
\textbf{Model} & \textbf{Strategy} & \textbf{rFID$\downarrow$}& \textbf{Speed$\downarrow$} \\
\midrule
NativeTok$_{64}$ & full & 6.50 & 1.53s \\
NativeTok$_{64}$ & reuse & 6.46 &1.15s \\
NativeTok$_{64}$ & reuse+fine tune & 6.22 & - \\
\bottomrule
\end{tabular*}

}
\caption{Training Strategies Comparison.}

\label{variants_table}
\end{subtable}
\hfill
\begin{subtable}[t]{0.28\textwidth}
\setlength{\tabcolsep}{1pt}
\renewcommand{\arraystretch}{1}
{\scriptsize
\begin{tabular*}{\linewidth}{@{\extracolsep{\fill}}lcc@{}}
\toprule
\textbf{model}  & \textbf{Sample/s} \\
\midrule
VQGAN &  233.02 \\
Titok-L-32 &  136.32\\
NativeTok$_{32}$ &  119.85\\
\bottomrule

\end{tabular*}
}
\caption{Encoding Speed Comparison.}

\label{speed}
\end{subtable}
\end{table*}
\label{sec:sub4.3}

\textbf{Comparison of different attention mechanisms:} In Table \ref{consolidated_table}, we compare different attention mechanisms used in NativeTok. We first conducted a comparative analysis of Titok$_{L-32}$, Titok$_{L(mask)-32}$ (which incorporates causal masks into its encoder), and NativeTok$_{32}$. This comparison further validates the structural efficiency of our proposed framework, especially when compared to traditional approaches that simply apply causal masks at the model input level, as the rFID significantly decreases from 12.99 and 12.95 to 11.19 under the same number of training steps.

\textbf{Comparison of Different Training Strategies.}
 Experiments here is designed to evaluate the efficiency of the proposed Hierarchical Native Training strategy and to identify the optimal training configuration which are conducted on two A800 GPUs with a per-GPU batch size of 64. In Table \ref{variants_table}, we conduct experiments on NativeTok$_{64}$ using three training strategies: (1) full-parameter training with random initialization, (2) full reuse of pretrained weights without modification (3) reuse training for 90\% of the steps followed by full-parameter fine-tuning in the remaining 10\%.
Results show that the Hierarchical Native Training strategy significantly improves training efficiency, reducing time per batch from 1.53s to 1.15s. At the same time, rFID improves from 6.50 to 6.46. After an additional 10\% of full-parameter fine-tuning, rFID is further reduced to 6.22, demonstrating the efficiency of this training strategy.

\textbf{Comparison of Encoding Speed.} As shown in Table~\ref{speed}, although NativeTok employs a longer MoCET sequence, its attention modeling is performed in a low-dimensional latent space. Owing to the $\mathcal{O}(n^2)$ complexity of transformers, this results in a moderate drop in encoding speed. However, the impact remains limited and does not significantly affect overall efficiency.

\subsection{Visualization}
\begin{figure}[H]
  \centering
  \includegraphics[width=0.8\linewidth]{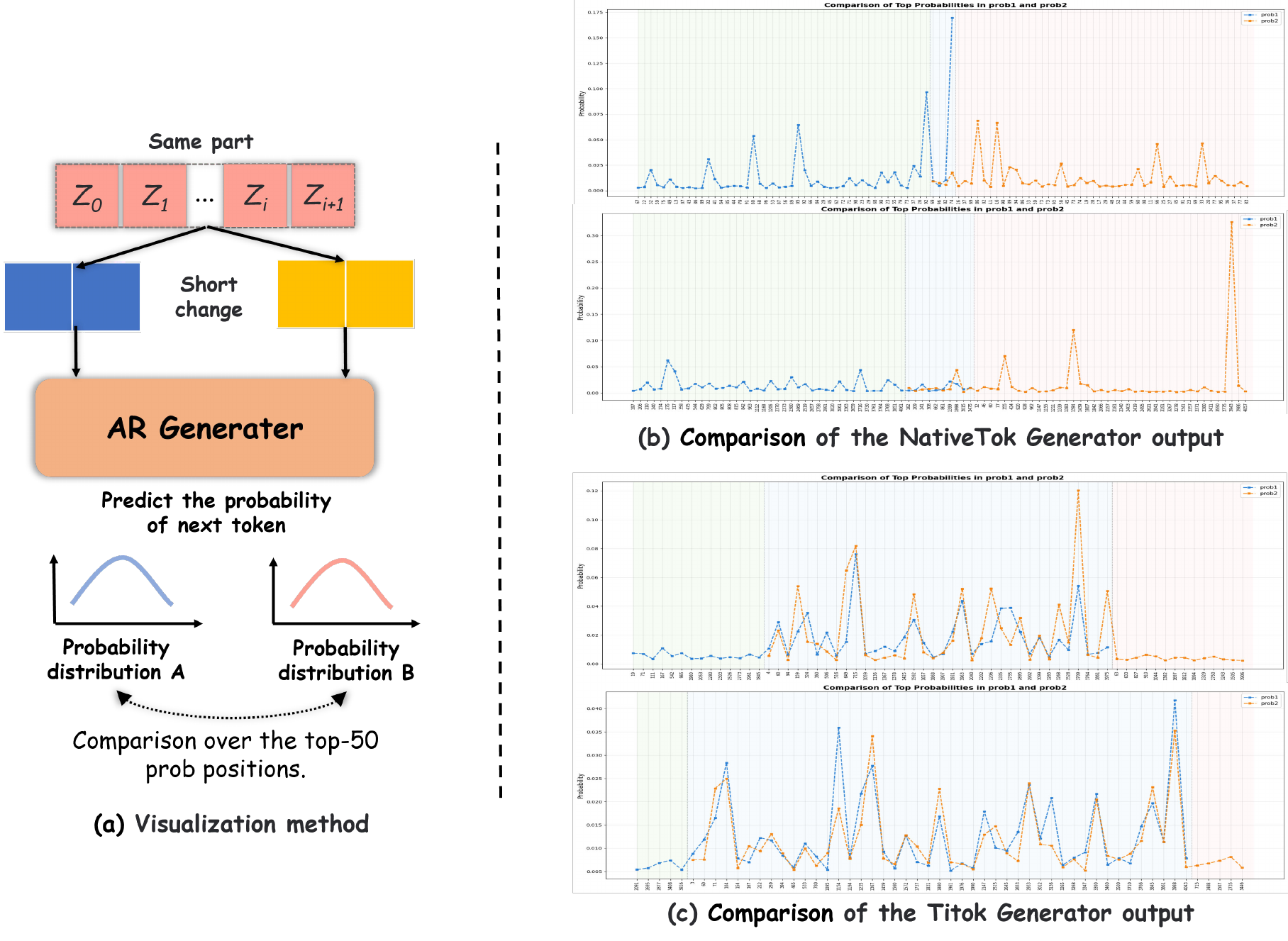}

  \caption{
    \textbf{Visualization: }  In the right-hand figure, the x-axis denotes the token position index, and the y-axis represents the corresponding probability values.
    When modifying adjacent token representations, we compare how the probability distribution of the next token changes accordingly. }
  \label{Visualization}
\end{figure}

In Fig.~\ref{Visualization}, we present visualization results that highlight the accuracy of NativeTok's token sequence to perturbations in preceding tokens. To investigate this, we first input a condition \( C \) into the generator and set two indices, \( i \) and \( j \). Tokens before position \( i \) are deterministically selected as the most probable choices, while tokens between positions \( i \) and \( j \) are sampled according to their probability distribution. We then visualize the probability distribution of the token generated at position \( j{+}1 \). Specifically, we extract the top-100 highest-probability positions from the output distributions of the same generator under two slightly different inputs. We categorize these into overlapping and non-overlapping subsets for comparison. Fig.~\ref{Visualization}(b) and Fig.~\ref{Visualization}(c) show the results for \textit{NativeTok$_{32}$} and \textit{TiTok$_{\text{L-32}}$}, respectively, under two settings: modifying 2 tokens (top row) and 4 tokens (bottom row). This observation is further supported by quantitative metrics showing a significant drop in the overlap of high-probability positions in the final token distribution, demonstrate that NativeTok yields a smaller overlapping region, suggesting that the generator models its token sequences with greater sensitivity and accuracy.

\section{Conclusion}
\label{sec:Conclusion}
We propose the concept of \textit{native visual tokenization} and introduce \textit{NativeTok}, a novel visual tokenizer that encodes images into ordered token sequences aligned with generation stage. By incorporating a native visual order and hierarchical training strategy, NativeTok bridges the gap between tokenization and generation, enabling improved image generation.

\clearpage
\section{Acknowledgment}
This research is supported by National Natural Science Foundation of China under Grant 623B2094.

\bibliographystyle{plainnat}
\bibliography{main}

\clearpage

\onecolumn
\section*{\centering\LARGE\bfseries NativeTok: Native Visual Tokenization for Improved Image Generation}
\section*{\centering Appendix}
\renewcommand{\thesection}{\Alph{section}}
\renewcommand{\thesubsection}{\Alph{section}.\arabic{subsection}}

\subsection{Training and Testing Protocols} \label{sec:intro}
\renewcommand{\arraystretch}{0.7}
\begin{table}[H]
\centering
\begin{tabular}{l|l}
\textbf{Configuration} & \textbf{Value} \\

\hline
\multicolumn{2}{c}{\textit{NativeTok$_{32}$}} \\
Meta Image Transformer layers & 18 \\
Meta Image Transformer dim & 1024 \\
MoCET layers & 32 \\
MoCET dim & 256 \\
decoder layers & 24 \\
decoder dim & 1024 \\
total parameters & 616M \\
training strategy & full param training \\
Trainable params rate & 100\% \\
speed & 63.44/s/gpu Batch \\
\hline
\multicolumn{2}{c}{\textit{NativeTok$_{64}$}} \\
Meta Image Transformer layers & 18 \\
Meta Image Transformer dim & 1024 \\
MoCET layers & 64 \\
MoCET dim & 256 \\
decoder layers & 24 \\
decoder dim & 1024 \\
total parameters & 666M \\
training strategy & HNT strategy \\
Trainable params rate & 56\% \\
speed & 54.14/s/gpu Batch \\
speed of full parameter fine-tune & 40.48/s/gpu Batch \\
\hline
\multicolumn{2}{c}{\textit{NativeTok$_{128}$}} \\

Meta Image Transformer layers & 18 \\
Meta Image Transformer dim & 1024 \\
MoCET layers & 128 \\
MoCET dim & 256 \\
decoder layers & 24 \\
decoder dim & 1024 \\
total parameters & 766M \\
training strategy & HNT strategy \\
Trainable params rate & 55\% \\
speed & 39.44/s/gpu Batch \\
speed of full parameter fine-tune & 30.75/s/gpu Batch \\
\hline
\multicolumn{2}{c}{\textit{Training Configuration}} \\

dateset & ImageNet1K \\
image resolution & 256\(\times\)256 \\
random crop& True\\
random flip& True\\
device & NVIDIA A800 $\times$ 4  \\
per gpu batch size & 64 \\
global batch size & 256 \\
optimizer & AdamW  \\
base learning rate & \(1\text{e}^{-4}\) \\
learning rate schedule & cosine \\
end learning rate & \(1\text{e}^{-5}\) \\
warmup steps & 5K \\
weight decay & \(1\text{e}^{-4}\) \\
optimizer momentum & \(\beta_1, \beta_2 = 0.9,\ 0.99\) \\
training steps & 500K \\
\hline
\end{tabular}
\caption{Configuration of NativeTok. } 
\label{tab:config}
\end{table}
\renewcommand{\arraystretch}{1} 
For image generation using \textbf{NativeTok}, we provide the configuration and training strategies across three model scales: \textit{NativeTok$_{32}$}, \textit{NativeTok$_{64}$}, and \textit{NativeTok$_{128}$}. All models are trained on ImageNet-1K at resolution \(256 \times 256\) with center cropping and horizontal flipping as the only augmentations.

\textbf{For \textit{NativeTok$_{32}$}}, we use a Meta Image Transformer with 18 layers and 1024 hidden dimensions, followed by a Mixture of Causal Expert Transformer composed of 32 layers with 256 hidden dimensions. The decoder has 24 layers and 1024 hidden dimensions. This model contains a total of 616M parameters and is trained with full parameter updates. The training throughput is 63.44 samples per second per GPU.

\textbf{For \textit{NativeTok$_{64}$}}, we increase the Mixture of Causal Expert Transformer to 64 layers while keeping other components unchanged. The model size grows to 666M parameters. We employ the Hierarchical Native Training (HNT) strategy over 450K training steps, resulting in only 56\% of parameters being updated during initial training. The training speed is 54.14 samples/s/GPU, and 40.48 samples/s/GPU for full-parameter fine-tuning.

\textbf{For \textit{NativeTok$_{128}$}}, we maintain the same architecture as \textit{NativeTok$_{64}$} but scale the training to a larger Mixture of Experts model with the same 64 expert layers. The parameter count rises to 766M. HNT strategy is used again with 55\% trainable parameters. Training throughput is 39.44 samples/s/GPU, dropping to 30.75 samples/s/GPU for full fine-tuning.

\textbf{Training configuration.} All models are trained using 4 NVIDIA A800 GPUs with a per-GPU batch size of 64 (global batch size 256). We use the AdamW optimizer with \(\beta_1 = 0.9\), \(\beta_2 = 0.99\), a base learning rate of \(1 \times 10^{-4}\), cosine decay schedule down to \(1 \times 10^{-5}\), and weight decay of \(1 \times 10^{-4}\). The warm-up phase lasts 5K steps, and the full training takes 500K steps.

\textbf{Hierarchical Native Training strategy.} For larger model variants (\textit{NativeTok$_{64}$} and \textit{NativeTok$_{128}$}), we use the Hierarchical Native Training (HNT) strategy, which gradually expands trainable parameter subsets to mitigate optimization difficulty in large-scale mixtures of experts.

We provide the complete configuration in table~\ref{tab:config}.

\subsection{Detailed Results of Preliminary Experiments} \label{sec:related}
\begin{table}[H]
\centering
\renewcommand{\arraystretch}{}
\setlength{\tabcolsep}{1pt}
{
\begin{tabular*}{\linewidth}{@{\extracolsep{\fill}}lcc@{}}
\toprule
\textbf{Model} & \textbf{Params$\downarrow$} & \textbf{rFID$\downarrow$} \\
\midrule
\multicolumn{3}{c}{stage1} \\
NativeTok$_{32}$ & 616M & 5.10 \\
NativeTok$_{64}$& 666M & 3.54 \\
NativeTok$_{128}$ & 766M & 2.86 \\
 \hline
 \multicolumn{3}{c}{add decoder finetune} \\

NativeTok$_{32}$ & 616M & 2.57 \\
NativeTok$_{64}$& 666M & 1.89 \\
NativeTok$_{128}$ & 766M & 1.19 \\
\bottomrule

\end{tabular*}
\caption{Detailed Performance of NativeTok.}
\label{generator}

}
\hfill

\hfill

\end{table}
\textbf{Detailed Performance of NativeTok.} We report the performance of different NativeTok variants across both training stages, as summarized in Table~\ref{generator}.

To quantitatively assess reconstruction quality, we measure rFID scores (lower is better) under two settings: \textit{Stage 1 only} and \textit{Stage 1 + decoder fine-tuning}. As shown in Table~\ref{generator}, increasing model capacity from NativeTok$_{32}$ to NativeTok$_{128}$ consistently enhances reconstruction quality. Under Stage 1 alone, the rFID decreases from 5.10 to 2.86 with increasing model scale. With additional decoder fine-tuning, the rFID further improves to 2.57, 1.89, and 1.19 for NativeTok$_{32}$, NativeTok$_{64}$, and NativeTok$_{128}$, respectively.

In addition, the parameter count reflects the impact of increasing the number of experts in the MoCET module. Although larger token counts result in more MoCET parameters, the increase is sublinear: every additional 32 tokens introduce only about an 8\% increase in parameters relative to the base model.

\subsection{Qualitative Visualization} \label{sec:method}

We provide more detailed quantitative metrics corresponding to the visualization section in the main paper in Table~\ref{overlap_comparison}. Specifically, we modify short token subsequences of varying lengths from 1 to 4 and compute the average top-100 overlap rate across 1,000 ImageNet classes. We compare the results of NativeTok and TiTok under the same setting. The results show that NativeTok consistently exhibits a lower overlap rate, indicating that the generator is more sensitive to token-level variations. This reflects a stronger alignment between the token representations and the generation behavior, resulting in more precise outputs.

\begin{figure*}[ht]
    \centering
    \includegraphics[width=0.8\linewidth]{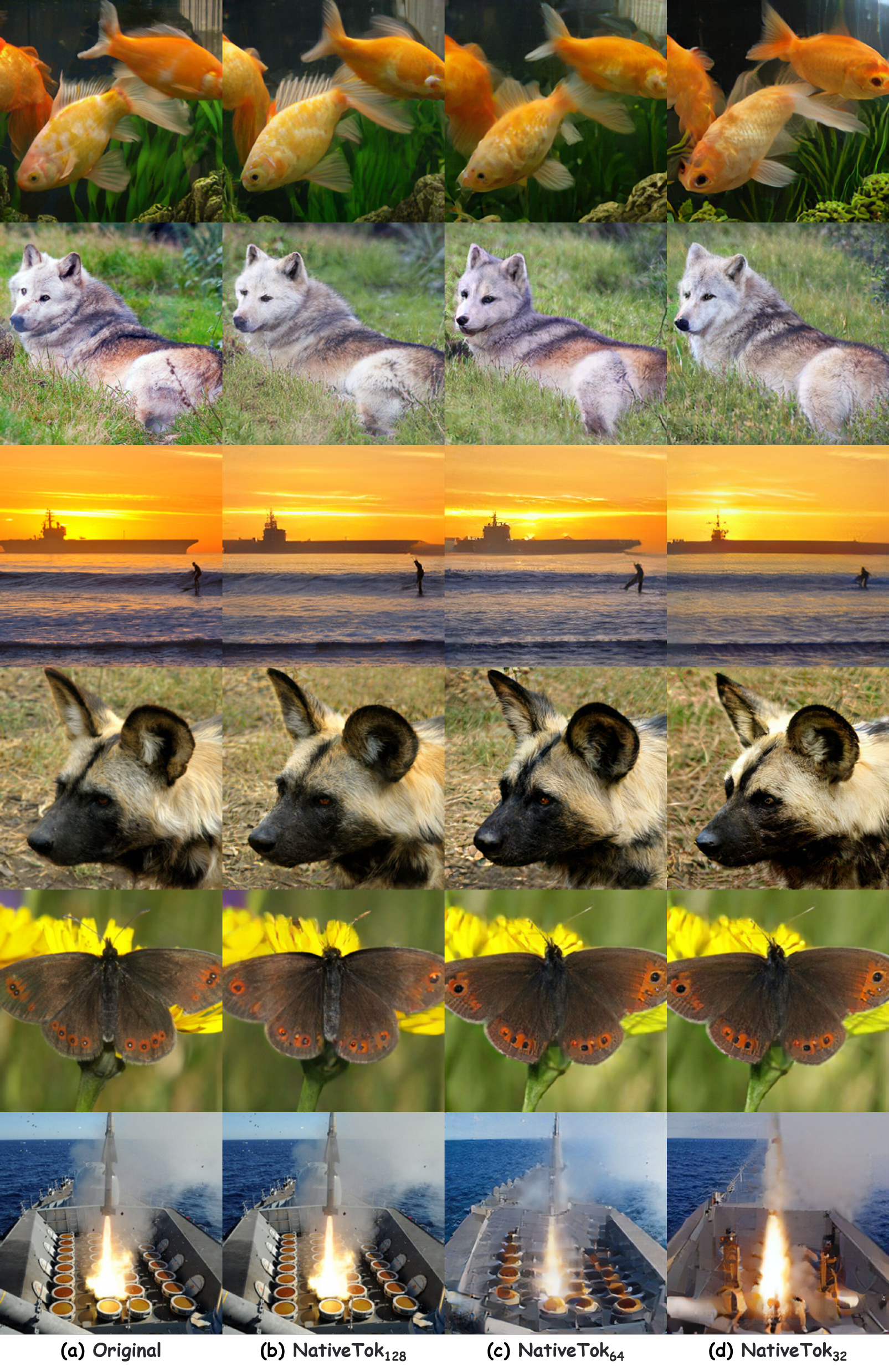}
    \caption{We provide a demonstration of reconstruction results. From left to right: the original image, reconstruction image by NativeTok$_{128}$, reconstruction image by NativeTok$_{64}$, and reconstruction image by NativeTok$_{32}$.}
    \label{fig:recon}
\end{figure*}

We provide several reconstruction visualizations in Figure~\ref{fig:recon}. From left to right: original image, NativeTok$_{128}$ reconstruction, NativeTok$_{64}$ reconstruction, and NativeTok$_{32}$ reconstruction.

\begin{table}[H]
\centering

\renewcommand{\arraystretch}{1}
\setlength{\tabcolsep}{12pt}
\begin{tabular}{>{\centering\arraybackslash}m{1.8cm} 
                >{\centering\arraybackslash}m{1.8cm} 
                >{\centering\arraybackslash}m{1.8cm}}
\toprule
\footnotesize\textbf{Number of Inconsistent Tokens} & 
\footnotesize\textbf{NativeTok Overlap Ratio} & 
\footnotesize\textbf{TiTok Overlap Ratio} \\
\midrule
1 & 56.26\% & 78.66\% \\
2 & 50.89\% & 75.78\% \\
3 & 48.34\% & 73.43\% \\
4 & 48.44\% & 68.37\% \\
\bottomrule
\end{tabular}
\caption{Comparison of Token Overlap Ratios between NativeTok and TiTok.}
\label{overlap_comparison}
\end{table}

\begin{figure*}[ht]
    \centering
    \includegraphics[width=1\linewidth]{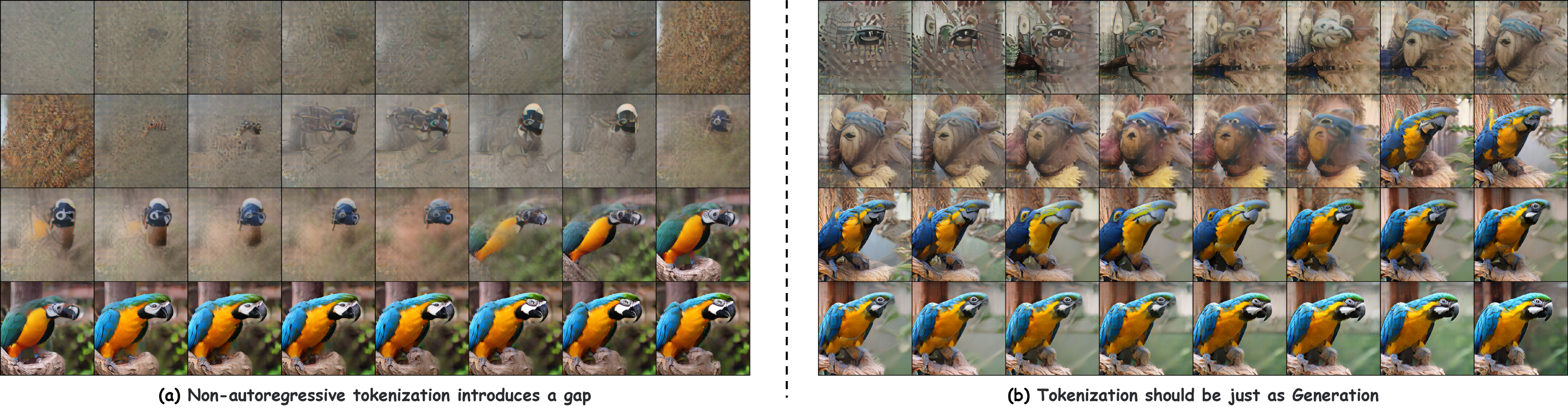}
    \caption{Reconstructions during the token generation process where the token count increases from 0 to 32.}
    \label{fig:process}
\end{figure*}
In Figure~\ref{fig:process}, we visualize the morphological changes of NativeTok$_{32}$ and Titok$_{32}$ reconstructions during the token generation process where the token count increases from 0 to 32. Our observations reveal that: Under Titok$_{32}$'s bidirectional attention mechanism, earlier tokens inherently contain information from subsequent tokens. In contrast, NativeTok$_{32}$ with unidirectional attention progressively incorporates novel visual information not present in previous tokens.

\begin{figure*}
    \centering
    \includegraphics[width=0.8\linewidth]{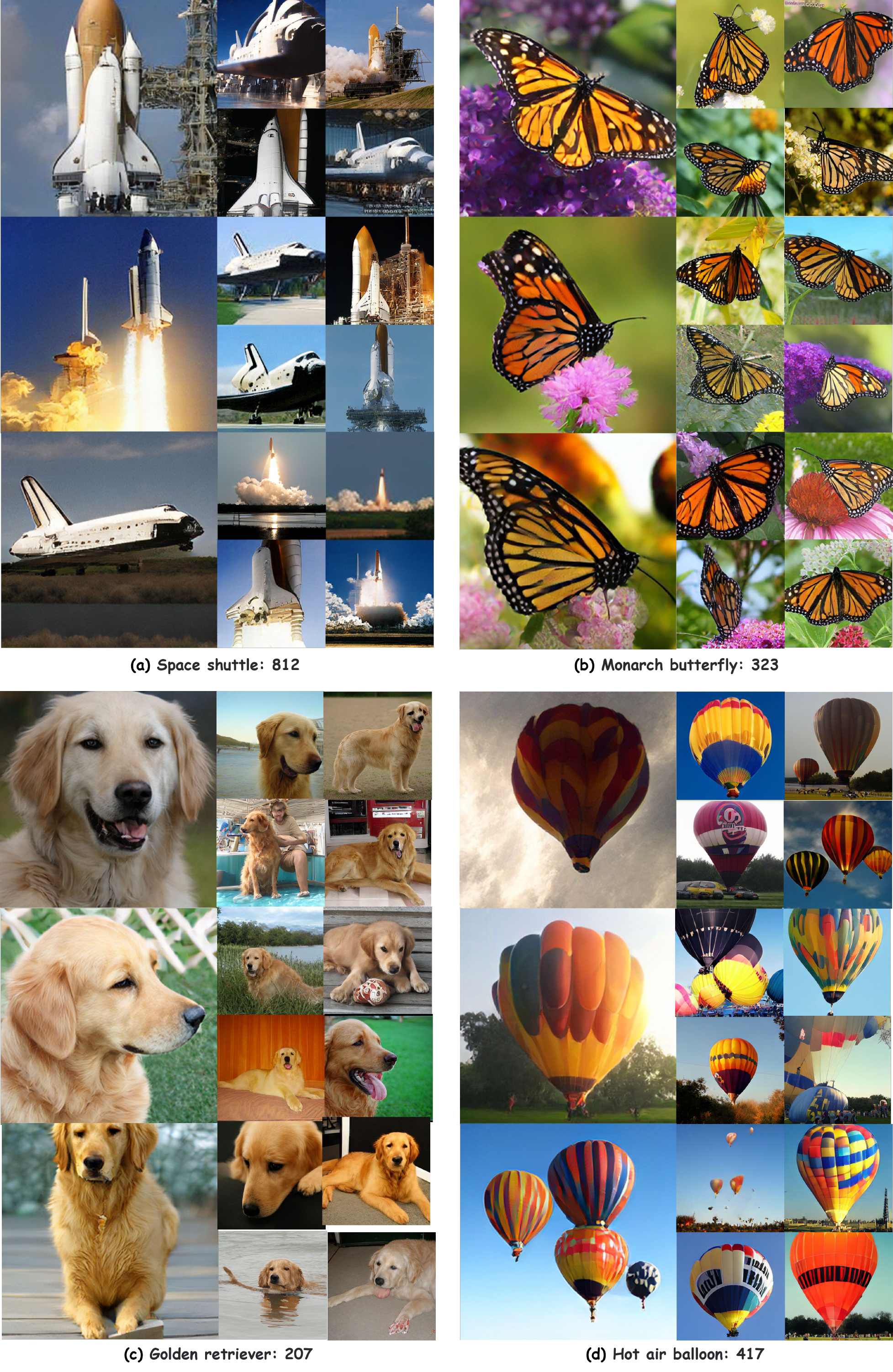}
    \caption{Generation examples}
    \label{fig:Generation}
\end{figure*}

We provide several qualitative examples of generated images in Figure~\ref{fig:Generation}.

\subsection{Limitations}
While our proposed method demonstrates promising performance in both reconstruction and generation tasks, several limitations remain to be addressed in future work.

\textbf{Two-stage Training Pipeline.}
Similar to previous methods such as TiTok, our framework still adopts a two-stage training scheme, where the image tokenizer is trained independently from the autoregressive generative model. Although we partially alleviate the gap between these stages through hierarchical causal training and decoder fine-tuning, a fully unified end-to-end training pipeline remains unexplored. Bridging this gap could further improve token quality and model coherence, especially in downstream autoregressive tasks.

\textbf{Limited Training Scale.}
Due to computational constraints and limited resources, we were only able to train the autoregressive generator using the smallest variant, \textit{NativeTok}${32}$. While this variant already shows clear advantages over its TiTok counterpart, it does not fully exploit the benefits of higher-capacity tokenizers like \textit{NativeTok}${128}$. As a result, our generation results may not reflect the upper bound of our proposed framework's potential. Future work will include large-scale training of the autoregressive model on more capable tokenizer variants, which is expected to yield further improvements.

\end{document}